\documentclass[10pt,twocolumn,letterpaper]{article}

\usepackage[pagenumbers]{cvpr} % To force page numbers, e.g. for an arXiv version

\usepackage[accsupp]{axessibility} % Improves PDF readability for those with visual impairments.
\usepackage{graphicx}
\usepackage{amsmath}
\usepackage{amssymb}
\usepackage{booktabs}
\usepackage{multirow}
\usepackage{xcolor}
\usepackage{bbm}

\usepackage[pagebackref,breaklinks,colorlinks]{hyperref}

\usepackage[capitalize]{cleveref}
\crefname{section}{Sec.}{Secs.}
\Crefname{section}{Section}{Sections}
\Crefname{table}{Table}{Tables}
\crefname{table}{Tab.}{Tabs.}

\def\cvprPaperID{9760} % *** Enter the CVPR Paper ID here
\def\confName{CVPR}
\def\confYear{2024}

\DeclareMathOperator*{\argmax}{argmax}

\begin{document}

\title{Retrieval-Augmented Open-Vocabulary Object Detection}

\newcommand\CoAuthorMark{\footnotemark[\arabic{footnote}]}
\author{
    Jooyeon Kim\textsuperscript{1,}\thanks{Equal contribution.} \quad
    Eulrang Cho\textsuperscript{2,}\protect\CoAuthorMark\hspace{0.15cm}\textsuperscript{,}\thanks{This work was done when she was working at Korea University.} \quad
    Sehyung Kim\textsuperscript{1} \quad
    Hyunwoo J. Kim\textsuperscript{1,}\thanks{Corresponding author.} \\
    \textsuperscript{1}Department of Computer Science and Engineering, Korea University~~~\textsuperscript{2}Samsung Research \\
    {\tt\small \{parang, shkim129, hyunwoojkim\}@korea.ac.kr~~~~~~ eulrang.cho@samsung.com}
}
\maketitle

%%%%%%%%% ABSTRACT
\begin{abstract}
    Open-vocabulary object detection (OVD) has been studied with Vision-Language Models (VLMs) to detect novel objects beyond the pre-trained categories.
    Previous approaches improve the generalization ability to expand the knowledge of the detector, using `positive' pseudo-labels with additional `class' names, e.g., sock, iPod, and alligator.
    To extend the previous methods in two aspects, 
    we propose Retrieval-Augmented Losses and visual Features (RALF).
    Our method retrieves related `negative' classes and augments loss functions.
    Also, visual features are augmented with `verbalized concepts' of classes, e.g., worn on the feet, handheld music player, and sharp teeth. 
    Specifically, RALF consists of two modules: Retrieval Augmented Losses (RAL) and Retrieval-Augmented visual Features (RAF). 
    RAL constitutes two losses reflecting the semantic similarity with negative vocabularies. 
    In addition, RAF augments visual features with the verbalized concepts from a large language model (LLM). 
    Our experiments demonstrate the effectiveness of RALF on COCO and LVIS benchmark datasets.
    We achieve improvement up to 3.4 box AP$_{50}^{\text{N}}$ on novel categories of the COCO dataset and 3.6 mask AP$_{\text{r}}$ gains on the LVIS dataset.  
    Code is available at \url{https://github.com/mlvlab/RALF}.
\end{abstract}

%%%%%%%%% BODY TEXT
\section{Introduction}
\label{sec:intro}
Open-vocabulary object detection (OVD) aims to detect objects belonging to open-set categories.
It is a challenging task because novel categories do not appear during training. 
Pre-trained vision-language models (VLMs) enable the detector to recognize base categories as well as novel categories via their zero-shot visual recognition ability learned from large-scale image-text pairs from the Internet.
For instance, CLIP~\cite{radford2021learning} and ALIGN~\cite{jia2021scaling} are widely used in OVD to classify unseen objects~\cite{gu2022open,wang2023object,lin2022learning}.

\begin{figure}[t]
    \centering
    \begin{subfigure}{0.97\linewidth}
        \includegraphics[width=1.0\linewidth]{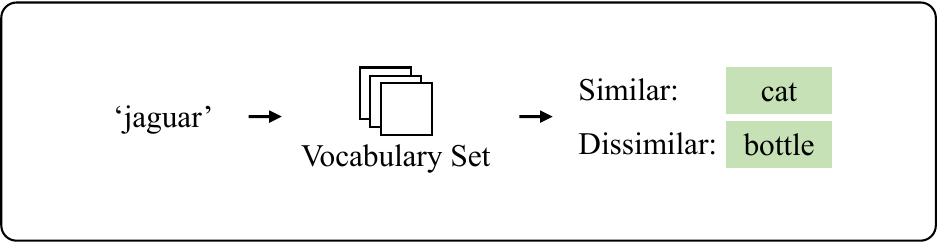}
        \caption{Retrieving hard (similar) and easy negative (dissimilar) classes.}
        \label{fig:cr_intro_vocab}
    \end{subfigure}
    \\[1em]
    \begin{subfigure}{0.97\linewidth}
        \includegraphics[width=1.0\linewidth]{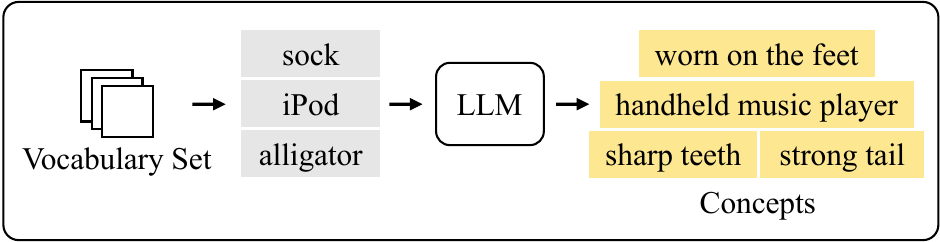}
        \caption{Verbalized concepts to augment visual features.}
        \label{fig:cr_intro_concept}
    \end{subfigure}    
    \caption{
       \textbf{Negative vocabularies and verbalized concepts from a large vocabulary set.}
        (a) Example of negative vocabularies that can be derived from a large vocabulary set. From the vocabulary set, `cat' and `bottle' can be retrieved as hard negative (similar) and easy negative (dissimilar) vocabulary, given the category `jaguar'.
        (b) Example of verbalized concepts that are generated from LLMs. The concepts of the objects provide more detailed information about the object, such as the attributes.}
    \label{fig:cr_intro}
\end{figure}

Knowledge distillation is one approach to transfer the knowledge of VLMs to the detector. It has been extensively explored in the literature~\cite{gu2022open,Hanoona2022Bridging,wang2023object,zhong2022regionclip,du2022learning}.
For more effective knowledge distillation, some works have endeavored to match words at the region level rather than at the image level to ensure better alignment~\cite{Hanoona2022Bridging,wang2023object,zhong2022regionclip}.
To improve generalization to novel categories, several studies employ pseudo-labeling to expand detector knowledge~\cite{zhong2022regionclip,gao2022open,zhao2022exploiting,cho2023open}. 
Pseudo-labels (positive classes) are generated for region proposals by matching their visual features with words from additional vocabulary sets or captions.
So, the additional vocabulary sets are constructed only focusing on `positive' classes.
We believe that more diverse vocabulary sets, such as `negative' classes and verbalized (visual) concepts with new techniques, will open the door to further improving OVD frameworks. 

To study the underexplored directions, in this paper
we propose methods utilizing `negative' classes and 
`verbalized concepts' to more effectively use vocabulary sets to enhance the generalization ability to novel categories. 
One approach involves retrieving vocabularies that have semantic relationships from large vocabulary sets when given a specific category. 
For example, given the category `jaguar', we retrieve `cat' and `bottle' as hard (similar) and easy (dissimilar) negative vocabulary, respectively as shown in \Cref{fig:cr_intro_vocab}.
Using these negative classes, we augment loss functions.
Another approach is generating verbalized concepts that describe the attributes (\eg, sharp teeth) of classes by LLMs, as depicted in \Cref{fig:cr_intro_concept}.
This information is useful in learning effective representations.

We present a novel framework, \textbf{R}etrieval-\textbf{A}ugmented  \textbf{L}osses and visual \textbf{F}eatures (RALF), which retrieves vocabularies and concepts from a large vocabulary set and augments losses and visual features.
RALF is composed of two parts: Retrieval-Augmented Losses (RAL) and Retrieval-Augmented visual Features (RAF).
Given the ground-truth label, we construct hard and easy negative vocabularies by retrieving from the vocabulary store based on similarity. 
Then, RAL optimizes the distance between the ground-truth label and pre-defined vocabularies with additional losses.
Additionally, we utilize a large language model (LLM) to obtain abundant information rather than word units. 
After generating descriptions about large vocabularies from LLM, we extract verbalized concept details that represent the characteristics of the objects and pile them in a concept store.
During inference time, RAF augments visual features with the verbalized concepts retrieved from the concept store.
Then the enhanced features are used in classification.
To validate the effectiveness of RALF, we conducted experiments on COCO and LVIS benchmarks. 
Overall, RALF improves the generalization ability of the detector.

Our contributions are threefold: 
\begin{itemize}
    \item  We present \textbf{RALF} that retrieves vocabularies and augments losses and visual features to improve the generalizability of open-vocabulary object detectors.
    \item \textbf{RAL} optimizes an embedding space by reflecting the distance between ground-truth labels and negative vocabularies from a large vocabulary set. And \textbf{RAF} augments visual features with  verbalized concepts relevant to visual attributes in images.
    \item With the method, we achieve 41.3 AP$^\text{N}_{50}$ in novel categories on the COCO benchmark and also reach 21.9 mask AP$_\text{r}$ in novel categories on LVIS. 
\end{itemize}
\begin{figure*}[t]
    \centering
    \begin{subfigure}{0.49\linewidth}
        \includegraphics[width=1.0\linewidth]{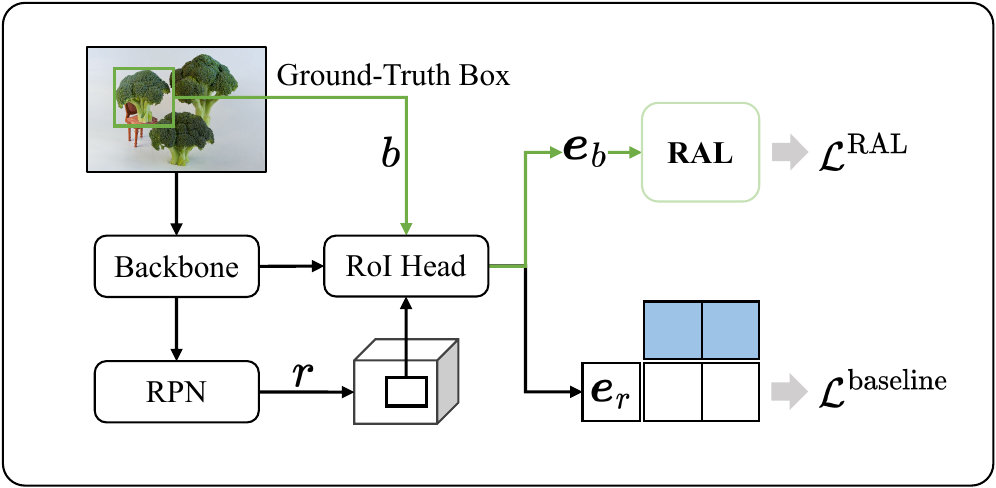}
        \caption{Training pipeline w/RAL.}
        \label{fig:cr_main_train_ral}
    \end{subfigure}
    \hfill
    \begin{subfigure}{0.49\linewidth}
        \includegraphics[width=1.0\linewidth]{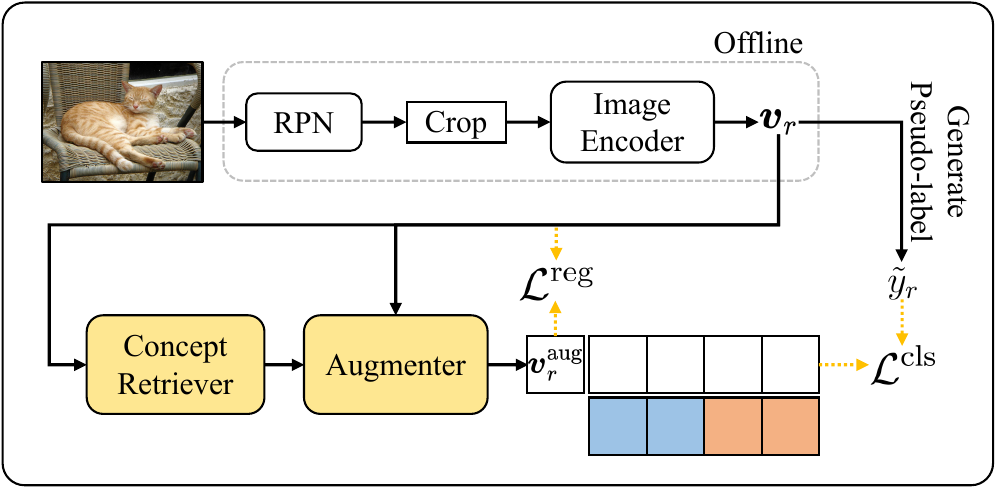}
        \caption{RAF training.}
        \label{fig:cr_main_train_raf}
    \end{subfigure} 
    \\[1em]
    \begin{subfigure}{1.0\linewidth}
        \centering
        \includegraphics[width=1.0\linewidth]{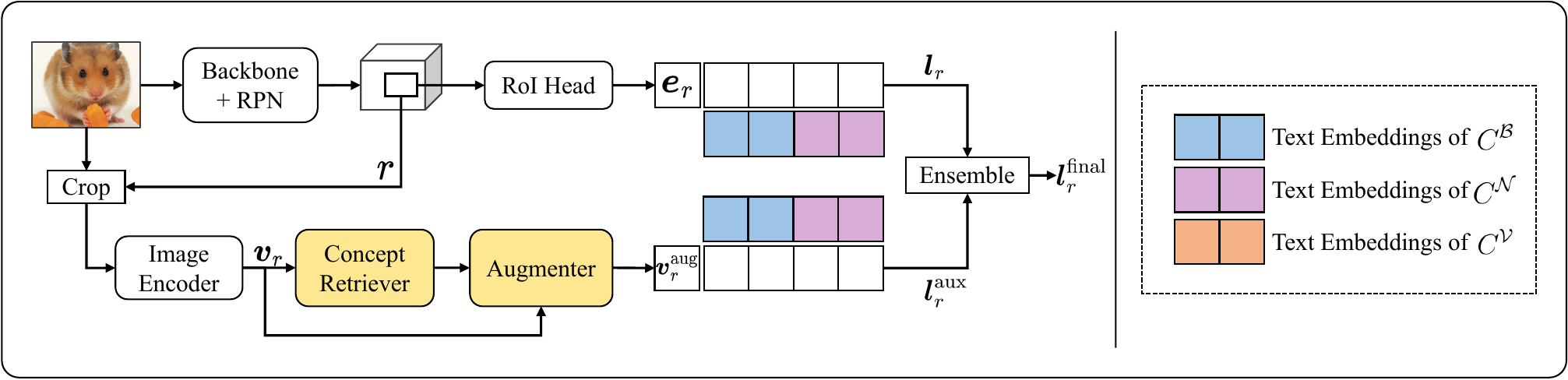}
        \caption{Inference pipeline w/RAF.}
        \label{fig:cr_main_test}
    \end{subfigure}
    \caption{\textbf{Overall pipeline of RALF.}
        (a) The first module, RAL, is utilized during detector training. Given a ground-truth box $b$, the ground-truth box embedding $\boldsymbol{e}_b$ is extracted and used to define $\mathcal{L}^\text{RAL}$, which is augmented with hard and easy negative vocabulary.
        The augmented loss $\mathcal{L}^\text{RAL}$ and the baseline loss $\mathcal{L}^\text{baseline}$ are employed together to train the detector.
        The illustration of the other branches (\eg, box regression, distillation, and mask prediction) is omitted in both training and inference pipelines.
        (b) The second module, RAF, augments visual features with verbalized concepts and is pre-trained before being used in the inference pipeline. 
        Augmented visual features $\boldsymbol{v}_r^\text{aug}$ are created through a process involving concept retriever and augmenter, using visual features $\boldsymbol{v}_r$ generated from object proposals in offline. RAF is trained with two losses ($\mathcal{L}^\text{cls}$ and $\mathcal{L}^\text{reg}$), utilizing $\boldsymbol{v}_r$, $\boldsymbol{v}_r^\text{aug}$, and $\tilde{y}_r$, which is the pseudo-label of visual feature.
        (c) During detector inference time, the trained RAF is utilized. Classification logits $\boldsymbol{l}_r$ trained by RAL and auxiliary logits $\boldsymbol{l}_r^\text{aux}$ influenced by RAF are computed with text embeddings of test categories. Then, the final logits $\boldsymbol{l}_r^\text{final}$ are determined through an ensemble of $\boldsymbol{l}_r$ and $\boldsymbol{l}_r^\text{aux}$.
    }
    \label{fig:cr_main}
\end{figure*}

\section{Related works}
\noindent\textbf{Pre-trained vision-language model.}
Pre-trained VLMs such as CLIP~\cite{radford2021learning} and ALIGN~\cite{jia2021scaling} are trained on large image-text pair datasets via contrastive learning for joint representation in visual and language modalities. Pre-trained VLMs consist of two encoders - image encoder and text encoder - extracting image embedding and text embedding, respectively. By applying contrastive learning, they can be aligned with each other in the same latent space. For this reason, pre-trained VLMs have superior generalizability and transferability to various downstream tasks. For example, CLIP~\cite{radford2021learning} which is pre-trained on extensive image-text pair data, shows impressive performance on zero-shot image classification tasks. Recently, thanks to the success of CLIP, there have been many studies for introducing VLMs to various downstream tasks such as image segmentation~\cite{zhou2022extract,rao2022denseclip,li2022languagedriven}, image generation~\cite{wang2022clip,ramesh2022hierarchical}, and object detection~\cite{gu2022open,wang2023object,Hanoona2022Bridging,wu2023aligning,lin2022learning}.

\noindent\textbf{Open-vocabulary object detection.}
Object detection task refers to a task that detects an object in a scene and classifies the detected object. A representative study, Fast R-CNN~\cite{girshick2015fast}, shows excellent object detection performance with CNN architecture. However, there is a limit to the object detection task that requires a lot of human cost for annotation. A zero-shot object detection approach was presented to determine whether a detector can detect a category that was not seen during learning. Recently, open-vocabulary object detection (OVD)~\cite{wu2023aligning,wang2023object,wu2023cora} has attracted attention. OVD evaluates the ability to predict novel categories by learning using additional caption data such as CC3M~\cite{sharma2018conceptual}. As pre-trained VLMs trained with large datasets perform well with zero-shot for various downstream tasks, diverse approaches to solve through pre-trained VLMs have been studied in OVD. ViLD~\cite{gu2022open}, the most representative study, uses knowledge from CLIP, one of the pre-trained VLMs, and learns the class-agnostic region proposal to perform well for unseen classes. Many follow-up studies have also been conducted on this, and Object-Centric-OVD~\cite{Hanoona2022Bridging} proposes an object-centric alignment method to solve the localization problem that occurs when CLIP is applied to OVD.

\noindent\textbf{Retrieval-augmentation.}
Retrieval augmentation was initially introduced in language generation tasks for parameter efficiency.
RAG~\cite{lewis2020retrieval} introduces generation models that combine parametric and non-parametric memory access. 
Recently, retrieval augmentation has been utilized in many vision tasks~\cite{long2022retrieval,xu2021texture,tseng2020retrievegan}.
RDMs~\cite{blattmann2022retrieval} suggest efficiently storing an image database and conditioning a relatively compact generative model.
EXTRA~\cite{ramos2023retrieval} proposes a retrieval-augmented image captioning model that enhances performance by leveraging cross-modal representations.
Unlike these methods employing retrieval augmentation in generation tasks, we applied retrieval augmentation to the OVD task for the first time in our knowledge.

\section{Method}
In this section, we proposed a new framework \textbf{RALF} that \textbf{R}etrieves information from a large vocabulary store and \textbf{A}ugments \textbf{L}osses and visual \textbf{F}eatures.
Before we delve into RALF, we briefly introduce the open-vocabulary object detection task in \Cref{subsec:prelim}. 
The overall pipeline of the proposed method is described in~\Cref{subsec:overview}.
Our method consists of two modules: 1) Retrieval-Augmented Losses (RAL) to train the object detector, and 2) Retrieval-Augmented visual Features (RAF) using 
the generated concepts by a large language model. 
The modules are presented in \Cref{subsec:ral} and \Cref{subsec:raf}, respectively. 

\subsection{Preliminaries}
\label{subsec:prelim}
Open-vocabulary object detection (OVD) is an advanced visual recognition task that extends the capability of traditional object detectors beyond pre-trained categories. 
OVD aims to \emph{localize} and \emph{classify} a wide range of objects, including the categories not encountered during training. 
In OVD, the pre-trained categories and unseen categories are called base categories $C^\mathcal{B}$ and novel categories $C^\mathcal{N}$, respectively.
In general, the approaches for OVD leverage a pre-trained region proposal method, \eg, Region Proposal Network (RPN), for category-agnostic (initial) localization. 
After the region proposal step, OVD methods utilize a pre-trained vision-language model to classify a wide range of categories in a zero-shot learning manner. 
Specifically, given region proposal $r$, the methods extract a region embedding $\boldsymbol{e}_r \in \mathbb{R}^d$ and compute the similarity or perform zero-shot classification with the text embeddings of category $\mathcal{T}(c) \in \mathbb{R}^d$, where $\mathcal{T}$ is the text encoder and $c$ is either a base or new category, \ie, $c \in  C^\mathcal{B}\cup C^\mathcal{N}$. 
Unless explicitly stated, vectors are row vectors in this paper.

\subsection{Overview of RALF}
\label{subsec:overview}
We present the overall pipeline of \textbf{R}etrieval-\textbf{A}ugmented \textbf{L}osses and visual \textbf{F}eatures \textbf{(RALF)}. 
As described in \Cref{fig:cr_main}, RALF consists of two modules: 1) \textbf{RAL} (\Cref{fig:cr_main_train_ral}), which retrieves negative vocabularies from the vocabulary store by semantic similarity, and enhances the loss function for training object detectors, 
and 2) \textbf{RAF} (\Cref{fig:cr_main_train_raf}), which augments visual features using verbalized concepts by a large language model given retrieved vocabulary. 

In RAL, we define two negative vocabulary sets $V_{y_{b}}^{\text{hard}}$ and  $V_{y_{b}}^{\text{easy}}$ to train the object detector. 
They are retrieved based on the semantic similarity score between the ground-truth class label $y_{b}$ and vocabularies from the external vocabulary store $C^{\mathcal{V}}$. 
In RAF, we extract the concepts associated with objects by LLM and retrieve them to augment visual feature $\boldsymbol{v}_{r} \in \mathbb{R}^d$ by using the augmenter $\mathcal{A}$ for better classification. 

Following the conventional OVD setup, our detector is trained with annotated bounding boxes of base categories $C^{\mathcal{B}}$.
As shown in \Cref{fig:cr_main_train_ral}, retrieval-augmented losses are additionally used for training. 
The total loss for training the object detector is defined as follows:
\begin{equation}
    \mathcal{L}^{\text{total}} = \mathcal{L}^\text{baseline} 
    +\mathcal{L}^\text{RAL},
\end{equation}
where $\mathcal{L}^\text{baseline}$ denotes the loss of the baseline and $\mathcal{L}^\text{RAL}$ is explained in \Cref{subsec:ral}.
We utilize the obtained conceptual information to train the RAF module independently.
As illustrated in \Cref{fig:cr_main_test},
RAF augments the visual feature in a plug-in manner during inference time.
Detailed explanations of RAL and RAF will be introduced in \Cref{subsec:ral} and \Cref{subsec:raf}, respectively.

\begin{figure}[t]
    \centering
    \includegraphics[width=0.95\linewidth]{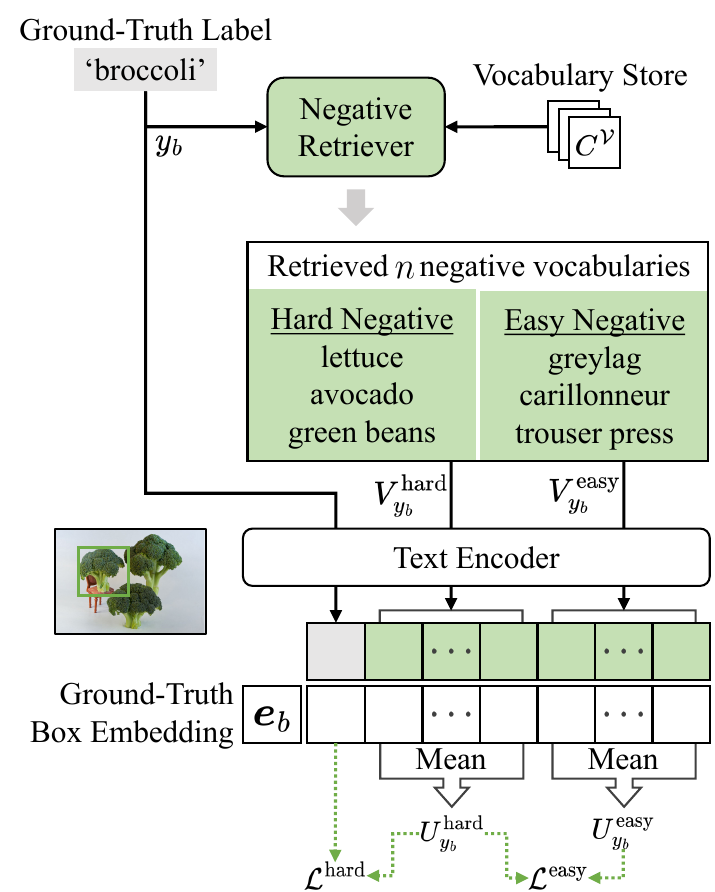}
    \caption{\textbf{RAL.}
        Given ground-truth class label $y_b$, negative retriever extracts hard negative vocabulary $V_{y_b}^\text{hard}$ and easy negative vocabulary $V_{y_b}^\text{easy}$ based on semantic similarity with $\mathcal{T}(y_b)$.
        To enhance the generalizability of the detector, two triplet losses (\ie hard negative loss $\mathcal{L}^\text{hard}$ and easy negative loss $\mathcal{L}^\text{easy}$) are augmented with $V_{y_b}^\text{hard}$, $V_{y_b}^\text{easy}$, and ground-truth box embedding $\boldsymbol{e}_b$.
    }
    \label{fig:cr_ral}
\end{figure}

\subsection{Retrieval-Augmented Losses}
\label{subsec:ral}
In our method, we introduce Retrieval-Augmented Losses (RAL), a novel framework utilizing a large vocabulary store to enhance 
the detector's generalization power across both base and novel object categories.
Our approach, as described in~\Cref{fig:cr_ral}, involves the creation of distinct negative vocabulary sets -- categorized as `hard' and `easy' -- based on
their similarities to the ground-truth category. 
RALF trains the detector with RAL, which is established with triplet losses between hard and easy negative vocabulary and ground-truth box.

\noindent\textbf{Negative retriever.}
To derive hard and easy negative vocabularies, we exploit a large vocabulary set containing a wider range of object classes.
First, we remove redundant categories from a vocabulary set and refine the vocabulary store $C^\mathcal{V}$ to prevent the possibility that the detector may see novel categories $C^\mathcal{N}$, \ie, $C^\mathcal{V} \cap C^\mathcal{N} = \varnothing$.
Then, we define hard negative vocabulary $V_{y_b}^{\text{hard}}$ and easy negative vocabulary $V_{y_b}^{\text{easy}}$ using relative similarity.
Specifically, given a ground-truth class label $y_b$, we obtain text embedding $\mathcal{T}(y_b) \in \mathbb{R}^d$.
The negative retriever samples hard and easy negative vocabularies from $C^\mathcal{V}$ with respect to cosine similarity between $\mathcal{T}(y_b)$ and $\mathcal{T}(C^\mathcal{V})$.
However, we observed that some vocabularies have constantly high (or low) similarity scores for any base category. In this case, the vocabularies are not useful to augment losses. To mitigate this issue, we adopt the rank variance sampling scheme, which filters out the vocabularies based on the variance of their rankings measured by similarity. Specifically, we first measure the similarity between a base category $c \in  C^\mathcal{B}$ and all the vocabularies in $C^\mathcal{V}$. Then, the ranking of $C^\mathcal{V}_i$ given $c$ is defined as:
\begin{equation}
    \underset{c \in C^\mathcal{B}}{\text{rank}_{c}(C_i^{\mathcal{V}})}  = 1+\sum_{ j \neq i} \mathbbm{1}[\mathcal{T}(c) \mathcal{T}(C^\mathcal{V}_j)^\top > \mathcal{T}(c) \mathcal{T}(C^\mathcal{V}_i)^\top]. 
\end{equation}
We finally compute the variance of $\text{rank}_{c}(C_i^{\mathcal{V}})$ using all $C^\mathcal{B}$. 
Vocabularies with a relatively low variance of rankings are removed.
Then, top-$m$ and bottom-$m$ vocabularies based on the similarity score are selected for each base category as hard and easy negative vocabularies, respectively. 
For each iteration of training, $n$ vocabularies are randomly selected among the $m$ vocabularies to augment losses. 
Further details of sampling schemes are discussed in the supplementary materials.

\noindent\textbf{Hard and easy negative losses.}
Hard and easy negative vocabularies are denoted as $V_{y_b}^{\text{hard}}$ and $V_{y_b}^{\text{easy}}$, respectively. 
$V_{y_b}^{\text{hard}}$ consists of the words similar to ground-truth $y_b$ class whereas $V_{y_b}^{\text{easy}}$ is the least similar words to $V_{y_b}^{\text{hard}}$.
Using the two sets of negative vocabularies with the triplet loss, we propose hard negative loss $\mathcal{L}^{\text{hard}}$ and easy negative loss $\mathcal{L}^{\text{easy}}$. 
Specifically, we first define the average cosine similarity with ground-truth embedding $\boldsymbol{e}_b$ as below:
\begin{align}
    & U_{y_b}^{\text{hard}} := \cfrac{1}{n} \; \mathbf{1} \mathcal{T}(V^\text{hard}_{y_b}) \boldsymbol{e}_b^\top, 
    \label{eq:avg_hard}
    \\
    & U_{y_b}^{\text{easy}} := \cfrac{1}{n} \; \mathbf{1} \mathcal{T}(V^\text{easy}_{y_b}) \boldsymbol{e}_b^\top,
    \label{eq:avg_easy}
\end{align}
where $\mathbf{1} \in \mathbb{R}^n$ given $n$ vocabularies.
Then, the hard negative loss $\mathcal{L}^{\text{hard}}$ and easy negative loss $\mathcal{L}^{\text{easy}}$ are defined as follows:
\begin{align}
    & \mathcal{L}^\text{hard} = \max \left( \lambda^\text{hard} U^\text{hard}_{y_b} - \mathcal{T}(y_b) \boldsymbol{e}_b^\top + \alpha^\text{hard}, 0 \right),
    \label{eq:loss_hard}
    \\
    & \mathcal{L}^\text{easy} = \max \left( \lambda^\text{easy} U^\text{easy}_{y_b} - U^\text{hard}_{y_b} + \alpha^\text{easy}, 0 \right),
    \label{eq:loss_easy}
\end{align}
where $\lambda^{\text{hard}}$ and $\lambda^{\text{easy}}$ are hyperparameters, and $\alpha^{\text{hard}}$ and $\alpha^{\text{easy}}$ denote margins.
To sum up, hard negative loss $\mathcal{L}^{\text{hard}}$ encourages $\boldsymbol{e}_b$ to have higher similarity with $y_b$ than $V_{y_b}^{\text{hard}}$.
Easy negative loss $\mathcal{L}^{\text{easy}}$ prompts $V_{y_b}^{\text{hard}}$ to exhibit relatively higher similarity to $y_b$ than $V_{y_b}^{\text{easy}}$.
The total loss $\mathcal{L}^{\text{RAL}}$ is computed with a summation of hard negative loss and easy negative loss as follows:
\begin{equation}
    \mathcal{L}^{\text{RAL}} = \beta^{\text{hard}}\mathcal{L}^{\text{hard}} + \beta^{\text{easy}}\mathcal{L}^{\text{easy}},
    \label{eq:raf_loss}
\end{equation}
where $\beta^{\text{hard}}$ and $\beta^{\text{easy}}$ are hyperparameters.

\subsection{Retrieval-Augmented visual Features}
\label{subsec:raf}
\begin{figure}[t]
    \centering
    \includegraphics[width=0.95\linewidth]{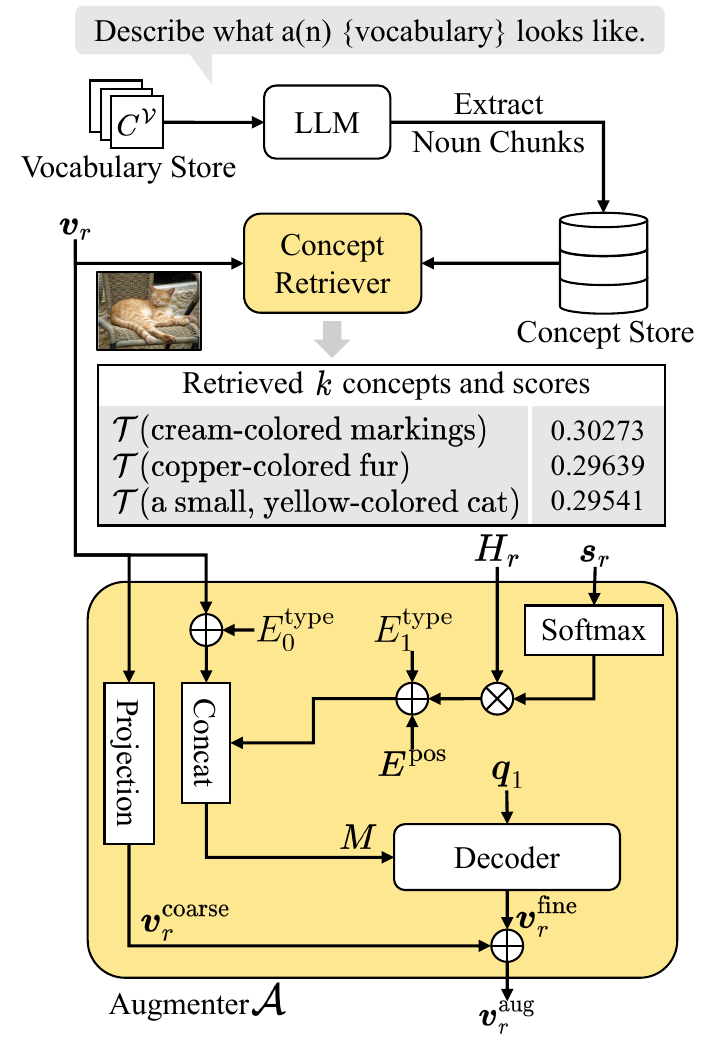}
    \caption{\textbf{RAF.}
        Verbalized concepts are generated by LLM with prompts and stored in the concept store. Given a visual feature $\boldsymbol{v}_r$, relevant concept embeddings $H_r$ and scores $\boldsymbol{s}_r$ are retrieved by the concept retriever. Then the augmenter $\mathcal{A}$ creates augmented visual feature $\boldsymbol{v}_r^\text{aug}$ with related verbalized concepts.
    }
    \label{fig:cr_raf}
\end{figure}

We introduce Retrieval-Augmented visual Features (RAF) that augments visual features using the \emph{verbalized concepts} of each object category, as depicted in \Cref{fig:cr_raf}.

\noindent\textbf{Concept store.}
The concept store consists of a set of characteristics describing the object (\eg, color, scale, shape, etc.).
The characteristics for $C^\mathcal{B} \cup C^\mathcal{V}$ are generated by a large language model (LLM) using `$\texttt{Describe}$ $\texttt{what a(n)}$ $\texttt{\{vocabulary\}}$ $\texttt{looks like.}$' prompt template, following~\cite{kaul2023multi}.
Note that $C^\mathcal{N}$ is not used to generate verbalized concepts.
We remove meaningless words such as prepositions from descriptions generated by LLM and store only meaningful noun chunks in the concept store.

\noindent\textbf{Concept retriever.}
Concepts for augmenting visual features are retrieved by the concept retriever.
Concept embeddings $H$ are obtained from the text encoder $\mathcal{T}$ with the concepts from the concept store.
Given visual feature $\boldsymbol{v}_r \in \mathbb{R}^d$, the concept retriever calculates the cosine similarity between concept embeddings $H$ and visual feature $\boldsymbol{v}_{r}$. 
Then, it returns the $k$ most relevant concept embeddings $H_{r} \in \mathbb{R}^{k \times d}$ and corresponding scores $\boldsymbol{s}_{r} \in \mathbb{R}^{k}$.

\noindent\textbf{Augmenter.}
We propose an augmenter $\mathcal{A}$ that augments visual feature $\boldsymbol{v}_{r}$ with the retrieved concepts.
Let $\boldsymbol{v}_r^\text{aug} \in \mathbb{R}^d$ denote a combination of $\boldsymbol{v}_{r}^{\text{coarse}} \in \mathbb{R}^d$ and $\boldsymbol{v}_{r}^{\text{fine}} \in \mathbb{R}^d$. 
$\boldsymbol{v}_{r}^{\text{coarse}}$ is calculated as:
\begin{equation}
    \boldsymbol{v}_r^\text{coarse} = \text{Proj}(\boldsymbol{v}_r),
    \label{eq:aug}
\end{equation}
where $\text{Proj}(\cdot)$ is a linear projection.
On the other hand, $\boldsymbol{v}_{r}^{\text{fine}}$ is enhanced with the retrieved concepts.
It is the final output of the decoder that has query embedding $\boldsymbol{q}_1 \in \mathbb{R}^d$ as query and $M \in \mathbb{R}^{(1 + k) \times d}$ as key and value. 
$M$ is computed as:
\begin{equation}
    M = (\boldsymbol{v}_r + E^\text{type}_0) \| (\text{diag}  (\text{softmax}(\boldsymbol{s}_r)) H_r + E^\text{pos} + E^\text{type}_1),
    \label{eq:M}
\end{equation}
where $\text{diag}(\cdot)$ denotes a diagonal matrix function and $\|$ is the concatenation function.
$E^\text{pos} \in \mathbb{R}^{k \times d}$ denotes positional embeddings to determine how many concepts will be utilized.
$E^\text{type}_0 \in \mathbb{R}^d$ and $E^\text{type}_1 \in \mathbb{R}^d$ are type embeddings to distinguish $\boldsymbol{v}_r$ between information from retrieved concepts. $E^\text{type}_1$ is replicated to $\mathbb{R}^{k \times d}$ in \cref{eq:M}.
The augmenter $\mathcal{A}$ consists of $L$ decoder layers. 
The operation in $l$-th decoder layer is as follows:
\begin{equation}
    \begin{split}
        \boldsymbol{q}^{\prime}_{l}&=\boldsymbol{q}_{l}+\text{CA}(\boldsymbol{q}_{l}, M, M), \\
        \boldsymbol{q}_{l+1}&= \boldsymbol{q}^{\prime}_{l}+\text{FFN}(\boldsymbol{q}^{\prime}_{l}), \\
        \boldsymbol{v}_{r}^{\text{fine}} &= \boldsymbol{q}_{L},
    \end{split}
\end{equation}
where $l \in \{1, \ldots, L\}$.
The term \text{CA} represents cross-attention operation, while \text{FFN} stands for feed-forward network.
Finally, the augmented visual feature $\boldsymbol{v}^\text{aug}_r$ is obtained by adding the coarse and fine feature of $\boldsymbol{v}_{r}$ as below:
\begin{equation}
    \boldsymbol{v}_r^\text{aug} = \boldsymbol{v}_{r}^{\text{coarse}} + \boldsymbol{v}_{r}^{\text{fine}}.
    \label{eq:aug_v_r}
\end{equation}
At test time, as depicted in \Cref{fig:cr_main_test} the augmented visual features and text embeddings of test categories are used to compute auxiliary logits $\boldsymbol{l}_r^\text{aux}$: 
\begin{equation}
    \boldsymbol{l}_r^\text{aux} = \boldsymbol{v}_r^\text{aug} {\mathcal{T}(C^\mathcal{B} \cup C^\mathcal{N})}^\top.
\end{equation}
Then the auxiliary logits are ensembled with $\boldsymbol{l}_r$ to compute the final logits $\boldsymbol{l}_r^\text{final}$ for the final classification of proposal $r$. The details of the logit ensemble are described in \Cref{sec:experiments}.

\noindent\textbf{Loss for RAF training.} 
We pre-trained RAF with visual features from region proposals.
For pre-training, we use classification loss $\mathcal{L}^{\text{cls}}$ and regularization loss $\mathcal{L}^{\text{reg}}$. 
We first define the pseudo-label of region proposal $r$ as $\tilde{y}_r$: 
\begin{equation}
    \tilde{y}_r = \argmax_{c \in (C^\mathcal{B} \cup C^\mathcal{V})} ( \mathcal{T}(c)\boldsymbol{v}_r^\top).
\end{equation}
Then, using the pseudo-labels, classification loss $\mathcal{L}^{\text{cls}}$ is defined as:
\begin{equation}
    \mathcal{L}^\text{cls} = \cfrac{1}{N}\sum_r \mathcal{L}^\text{ce}(\boldsymbol{v}^\text{aug}_r {\mathcal{T}(C^\mathcal{B} \cup C^\mathcal{V})}^\top, \tilde{y}_r), 
    \label{eq:cls_loss}
\end{equation}
where $N$ is the number of proposals per image and $\mathcal{L}^\text{ce}$ is the cross entropy loss.
We encourage augmented visual feature $\boldsymbol{v}_r^{\text{aug}}$ to similar to the original visual feature $\boldsymbol{v}_r$ using regularization loss $\mathcal{L}^{\text{reg}}$ defined as below:
\begin{equation}
    \mathcal{L}^\text{reg} = \cfrac{1}{N}\sum_r (\boldsymbol{v}^\text{aug}_r - \boldsymbol{v}_r)^2.
    \label{eq:reg_loss}
\end{equation}
Finally, the total loss $\mathcal{L}^{\text{RAF}}$ for RAF training is the combination of~\cref{eq:cls_loss} and~\cref{eq:reg_loss} as follows:
\begin{equation}
    \mathcal{L}^{\text{RAF}} = \beta^{\text{cls}}\mathcal{L}^{\text{cls}} + \beta^{\text{reg}}\mathcal{L}^{\text{reg}},
\end{equation}
where $\beta^{\text{cls}}$ and $\beta^{\text{reg}}$ are hyperparameters.

\section{Experiments}
\label{sec:experiments}
In this section, we briefly discuss the experimental setup, including datasets and implementation details.
Next, we evaluate the performance of RALF compared to various baselines and investigate RALF with further analysis. 

\noindent\textbf{Datasets.}
We evaluate RALF on two public benchmarks, COCO~\cite{lin2014microsoft} and LVIS~\cite{gupta2019lvis}. 
In the open-vocabulary object detection setting, COCO dataset is split into 48 base categories and 17 novel categories following OVR-CNN~\cite{zareian2021open}. 
It includes 118k images, which separated 107,761 images for training and 4,836 images for validation. 
Referring to ViLD~\cite{gu2022open}, we divide the LVIS dataset into 866 base categories and 337 novel categories. 
For both benchmarks, we use base categories during training, yet novel categories are also validated for inference. 
We adopt the mean average precision (mAP) as the evaluation metric. 
We report $\text{AP}_{\text{50}}$, ${\text{AP}}^{\text{B}}_{\text{50}}$, and ${\text{AP}}^{\text{N}}_{\text{50}}$ for COCO and $\text{AP}_{\text{r}}$, $\text{AP}_{\text{c}}$, ${\text{AP}}_{\text{f}}$, and ${\text{AP}}$ for LVIS. 
Note that instance segmentation (mAP) results are reported on LVIS.
To generate the vocabulary store, we adopt V3Det~\cite{wang2023v3det} as a vocabulary set. 

\noindent\textbf{Implementation details.}
We implement RALF with pre-trained CLIP~\cite{radford2021learning} with ViT-B/32 image encoder backbone and text encoder from the official repository. 
Note that we freeze all parameters in image and text encoders during training. 
Additionally, we use GPT-3~\cite{brown2020language} DaVinci-002 to generate descriptions from RAF as a large language model. 
We use Faster R-CNN~\cite{ren2015faster} with ResNet-50~\cite{he2016deep} backbone and Mask R-CNN~\cite{he2017mask} with ResNet-50 FPN backbone for COCO and LVIS, respectively. 
For training RAF, we use region proposals provided from OADP~\cite{wang2023object}. 
We set our baselines as OADP~\cite{wang2023object}, Object-Centric-OVD~\cite{Hanoona2022Bridging}, and DetPro~\cite{du2022learning}. 
As RALF enhances the generalizability power via a plug-in manner, we construct RALF into baselines. 
In all experiments, we train and evaluate on NVIDIA RTX-3090 4 GPUs. 
More implementation details, including hyperparameters, are discussed in the supplementary materials.

\noindent\textbf{Logit ensemble in RAF.}
Since each baseline has a different range of logits, the final logits $\boldsymbol{l}_r^\text{final}$ are calculated as below:
\begin{equation}
    \boldsymbol{l}_r^\text{final} = 
    \begin{cases}
        \boldsymbol{l}_r + \sigma (\boldsymbol{l}_r^\text{aux}) & \text{if Object-Centric-OVD}, \\
        \boldsymbol{l}_r * (\sigma (\boldsymbol{l}_r^\text{aux}) - 0.25) & \text{if DetPro}, \\
        \boldsymbol{l}_r + \boldsymbol{l}_r^\text{aux} & \text{if OADP}, \\
    \end{cases}
\end{equation}
where $\sigma(\cdot)$ denotes the sigmoid function.
Without using all values of auxiliary logits when adding to $\boldsymbol{l}_r$, only top-1 figures on COCO and top-10 or 20 figures on LVIS are utilized, considering the number of test categories.

\subsection{Main results}
We evaluate RALF on COCO and LVIS benchmarks in the open-vocabulary object detection (OVD) setting and compare with various baselines. Overall results are reported in ~\Cref{tab:cr_coco} and ~\Cref{tab:cr_lvis}.

\noindent{\textbf{COCO benchmark.}}
As reported in ~\Cref{tab:cr_coco}, RALF shows great performance enhancement when plugged into baselines for all evaluation metrics. 
RALF significantly improved 4.7 $\text{AP}^\text{N}_\text{50}$ when plugged into Object-Centric-OVD~\cite{Hanoona2022Bridging} and achieved state-of-the-art results. 
Moreover, RALF surpasses Object-Centric-OVD for not only novel categories but also base and all categories, with 0.3 $\text{AP}^\text{B}_\text{50}$ and 1.5 $\text{AP}_\text{50}$ improvements, respectively.
Like the inclination observed in the Object-Centric-OVD, implementing RALF on OADP~\cite{wang2023object} demonstrates remarkable effectiveness. 
The results show performance gains for the novel, base, and all categories by 3.4 $\text{AP}^\text{N}_\text{50}$, 1.2 $\text{AP}^\text{B}_\text{50}$, and 1.8 $\text{AP}_\text{50}$, correspondingly. 
\begin{table}[t]
  \centering
  \scalebox{0.98}{
      \begin{tabular}{l | c | c c c}
        \toprule
        Method & AP$^\text{N}_{50}$ & AP$^\text{B}_{50}$ & AP$_{50}$ \\
        \midrule
        ViLD~\cite{gu2022open} & 27.6 & \color{gray}{59.9} & \color{gray}{51.2} \\
        PB-OVD~\cite{gao2022open} & 29.1 & \color{gray}{44.4} & \color{gray}{40.4} \\
        OV-DETR~\cite{zang2022open} & 29.4 & \color{gray}{61.0} & \color{gray}{52.7} \\
        VL-Det~\cite{lin2022learning} & 32.0 & \color{gray}{50.6} & \color{gray}{45.8} \\
        MEDet~\cite{chen2022open} & 32.6 & \color{gray}{53.5} & \color{gray}{48.0} \\
        BARON~\cite{wu2023aligning} & 34.0 & \color{gray}{60.4} & \color{gray}{53.5} \\
        \midrule\midrule
        OADP~\cite{wang2023object} & 30.0 & \color{gray}{53.3} & \color{gray}{47.2} \\
        OADP + RALF & \textbf{33.4} & \color{gray}{\textbf{54.5}} & \color{gray}{\textbf{49.0}} \\
        \midrule
        Object-Centric-OVD~\cite{Hanoona2022Bridging} & 36.6 & \color{gray}{54.0} & \color{gray}{49.4}\\
        Object-Centric-OVD + RALF & \textbf{41.3} & \color{gray}{\textbf{54.3}} & \color{gray}{\textbf{50.9}} \\
        \bottomrule
      \end{tabular}
  }
  \caption{\textbf{Results of OVD on COCO.}}
  \label{tab:cr_coco}
\end{table}

\begin{table}[t]
  \centering
  \scalebox{0.91}{
      \begin{tabular}{l|c|ccc}
        \toprule
        Method & AP$_\text{r}$ & AP$_\text{c}$ & AP$_\text{f}$ & AP \\
        \midrule
        ViLD~\cite{gu2022open} & 16.1 & \color{gray}{20.0} & \color{gray}{28.3} & \color{gray}{22.5} \\
        OV-DETR~\cite{zang2022open} & 17.4 & \color{gray}{25.0} & \color{gray}{32.5} & \color{gray}{26.6} \\
        BARON~\cite{wu2023aligning} & 18.0 & \color{gray}{24.4} & \color{gray}{28.9} & \color{gray}{25.1} \\
        \midrule\midrule
        Object-Centric-OVD~\cite{Hanoona2022Bridging} & 17.1 & \color{gray}{21.4} & \color{gray}{26.7} & \color{gray}{22.8} \\
        Object-Centric-OVD + RALF & \textbf{18.5} & \color{gray}{21.0} & \color{gray}{26.3} & \color{gray}{22.6} \\
        \midrule
        DetPro~\cite{du2022learning} & 19.8 & \color{gray}{25.6} & \color{gray}{28.9} & \color{gray}{25.9} \\
    DetPro + RALF & \textbf{21.1} & \color{gray}{\textbf{25.7}} & \color{gray}{\textbf{29.2}} & \color{gray}{\textbf{26.3}} \\
        \midrule
        OADP~\cite{wang2023object} & 19.9 & \color{gray}{26.0} & \color{gray}{28.7} & \color{gray}{26.0} \\
        OADP + RALF & \textbf{21.9} & \color{gray}{\textbf{26.2}} & \color{gray}{\textbf{29.1}} & \color{gray}{\textbf{26.6}} \\
        \bottomrule
      \end{tabular}
    }
  \caption{\textbf{Results of OVD on LVIS.}}
  \label{tab:cr_lvis}
\end{table}

\noindent{\textbf{LVIS benchmark.}}
To verify that our method improves performance in various cases, we experimented on the LVIS benchmark and added another baseline -- DetPro~\cite{du2022learning}. 
The results are noted in~\Cref{tab:cr_lvis}. 
Overall, RALF improves detection ability on novel categories for all baselines by up to 2.0 $\text{AP}_{\text{r}}$. 
Although Object-Centric-OVD~\cite{Hanoona2022Bridging} slightly decreases except $\text{AP}_{\text{r}}$, RALF improves all metrics on DetPro and OADP~\cite{wang2023object}.
To summarize, the results imply that RALF improves generalizability. 

\subsection{Ablation study}
As discussed in \Cref{subsec:overview}, RALF consists of two modules -- RAL and RAF. 
We conducted ablation studies on RAL and RAF to verify the effectiveness of each module on COCO and LVIS benchmarks. 

\noindent\textbf{Effectiveness of RAL.}
We demonstrate the results of RAL in \Cref{tab:cr_ablation_coco} and \Cref{tab:cr_ablation_lvis}. 
From the results, RAL improves performance gain in all baselines. 
RAL shows an improvement of up to 1.3 $\text{AP}_{\text{50}}^{\text{N}}$ and at least 1.0 $\text{AP}_{\text{50}}^{\text{N}}$ on COCO. 
Interestingly, RAL demonstrates exceptional generalizability, specifically within the COCO base categories, exhibiting no performance declines. 
While there are marginal decreases in $\text{AP}_{\text{c}}$, $\text{AP}_{\text{f}}$, and $\text{AP}$ on LVIS, the performance of novel categories is notably increased up to 3.2 $\text{AP}_{\text{r}}$.

\begin{table}[t]
  \centering
  \scalebox{0.85}{
      \begin{tabular}{l|cc|cccc}
        \toprule
        Method & RAL & RAF & AP$^\text{N}_{50}$ & AP$^\text{B}_{50}$ & AP$_{50}$ \\
        \midrule
        \multirow{4}{*}{OADP$^\dagger$~\cite{wang2023object}} &  & & 30.0 & \color{gray}{54.5} & \color{gray}{48.1} \\
        & & \checkmark & 31.5 & \color{gray}{54.3} & \color{gray}{48.3} \\
        & \checkmark &  & 31.3 & \color{gray}{54.5} & \color{gray}{48.4} \\
        & \checkmark & \checkmark & 33.4 & \color{gray}{54.5} & \color{gray}{49.0} \\
        \midrule
        \multirow{4}{*}{Object-Centric-OVD$^\dagger$~\cite{Hanoona2022Bridging}} & & & 40.0 & \color{gray}{53.8} & \color{gray}{50.2} \\
         & & \checkmark & 40.8 & \color{gray}{54.0} & \color{gray}{50.5} \\
        & \checkmark &  & 41.0 & \color{gray}{54.1} & \color{gray}{50.7} \\
        & \checkmark & \checkmark & 41.3 & \color{gray}{54.3} & \color{gray}{50.9} \\
        \bottomrule
      \end{tabular}
  }
  \caption{\textbf{Ablation study of RALF on COCO dataset.} $\dagger$ denotes reproduce result.}
  \label{tab:cr_ablation_coco}
\end{table}

\begin{table}[t]
  \centering
  \scalebox{0.78}{
      \begin{tabular}{l|cc|cccc}
        \toprule
        Method & RAL & RAF & AP$_\text{r}$ & AP$_\text{c}$ & AP$_\text{f}$ & AP \\
        \midrule
        \multirow{4}{*}{Object-Centric-OVD$^\dagger$~\cite{Hanoona2022Bridging}} & &  & 18.1 & \color{gray}{21.4} & \color{gray}{26.6} & \color{gray}{22.8} \\
        & & \checkmark & 19.5 & \color{gray}{21.4} & \color{gray}{26.6} & \color{gray}{23.1} \\
        & \checkmark &  & 18.5 & \color{gray}{20.8} & \color{gray}{26.3} & \color{gray}{22.6} \\
        & \checkmark & \checkmark & 18.5 & \color{gray}{21.0} & \color{gray}{26.3} & \color{gray}{22.6} \\
        \midrule
        \multirow{4}{*}{DetPro$^\dagger$~\cite{du2022learning}}& & & 19.9 & \color{gray}{25.8} & \color{gray}{29.2} & \color{gray}{26.1} \\
        & & \checkmark & 20.5 & \color{gray}{25.8} & \color{gray}{29.2} & \color{gray}{26.2} \\
        & \checkmark &  & 21.3 & \color{gray}{25.7} & \color{gray}{29.2} & \color{gray}{26.3} \\
        & \checkmark & \checkmark & 21.1 & \color{gray}{25.7} & \color{gray}{29.2} & \color{gray}{26.3} \\
        \midrule
        \multirow{4}{*}{OADP$^\dagger$~\cite{wang2023object}}& & & 18.3 & \color{gray}{26.2} & \color{gray}{28.9} & \color{gray}{25.9} \\
        & & \checkmark & 19.3 & \color{gray}{26.2} & \color{gray}{28.9} & \color{gray}{26.1} \\
        & \checkmark &  & 21.5 & \color{gray}{26.1} & \color{gray}{29.1} & \color{gray}{26.5} \\
        & \checkmark & \checkmark & 21.9 & \color{gray}{26.2} & \color{gray}{29.1} & \color{gray}{26.6} \\
        \bottomrule
      \end{tabular}
  }
  \caption{\textbf{Ablation study of RALF on LVIS dataset.} $\dagger$ denotes reproduce result.}
  \label{tab:cr_ablation_lvis}
\end{table}

\begin{figure*}[t]
    \centering
    \rotatebox{90}{\makebox[7.5em][r]{(a) OADP}}
    \begin{subfigure}{0.23\textwidth}
        \includegraphics[trim={0 1em 0 20em},clip,width=\textwidth]{Figures/Materials/qual_1.png}
    \end{subfigure}
    \hspace{0.01\textwidth}
    \begin{subfigure}{0.23\textwidth}
        \includegraphics[width=\textwidth]{Figures/Materials/qual_3.png}
    \end{subfigure}
    \hspace{0.01\textwidth}
    \begin{subfigure}{0.23\textwidth}
        \includegraphics[width=\textwidth]{Figures/Materials/qual_5.png}
    \end{subfigure}
    \hspace{0.01\textwidth}
    \begin{subfigure}{0.23\textwidth}
        \includegraphics[trim={0 0 0 5em},clip,width=\textwidth]{Figures/Materials/qual_7.png}
    \end{subfigure}
    \\[1em]
    \rotatebox{90}{\makebox[9.5em][r]{(b) OADP + RALF}}
    \begin{subfigure}{0.23\textwidth}
        \includegraphics[trim={0 1em 0 20em},clip,width=\textwidth]{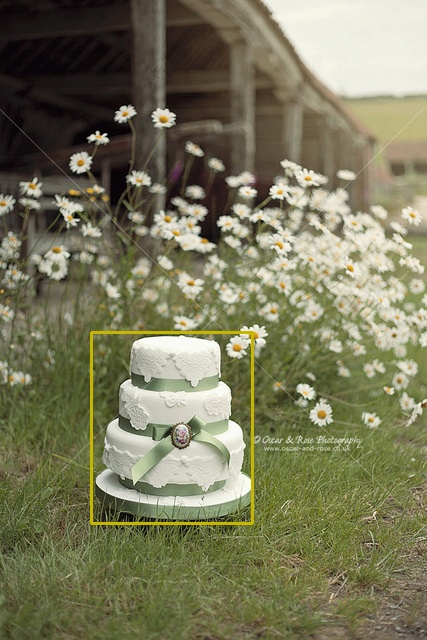}
    \end{subfigure}
    \hspace{0.01\textwidth}
    \begin{subfigure}{0.23\textwidth}
        \includegraphics[width=\textwidth]{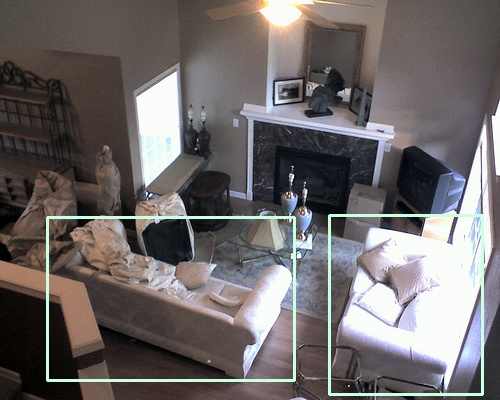}
    \end{subfigure}
    \hspace{0.01\textwidth}
    \begin{subfigure}{0.23\textwidth}
        \includegraphics[width=\textwidth]{Figures/Materials/qual_6.png}
    \end{subfigure}
    \hspace{0.01\textwidth}
    \begin{subfigure}{0.23\textwidth}
        \includegraphics[trim={0 0 0 5em},clip,width=\textwidth]{Figures/Materials/qual_8.png}
    \end{subfigure}
    \caption{\textbf{Qualitative results on COCO.} Results of bounding box predictions on novel categories for (a) OADP and (b) OADP+RALF.}
    \label{fig:qualitative}
\end{figure*}

\noindent\textbf{Effectiveness of RAF.}
We plug RAF into several baselines and evaluate performance to verify whether RAF is effective. 
~\Cref{tab:cr_ablation_coco} and~\Cref{tab:cr_ablation_lvis} show the performance of RAF on COCO and LVIS, respectively. 
There are notable improvements in overall baselines. 
In particular, RAF improves the prediction ability for novel categories of OADP and Object-Centric-OVD by 1.5 $\text{AP}_{\text{50}}^{\text{N}}$ and 0.8 $\text{AP}_{\text{50}}^{\text{N}}$ on COCO, respectively. 
Without performance decreasing in base categories, the performance of novel categories improved by up to 1.4 $\text{AP}_{\text{r}}$ on LVIS.

From the results in ~\Cref{tab:cr_ablation_coco} and ~\Cref{tab:cr_ablation_lvis}, irrespective of the baselines, RAL and RAF enhance the prediction of novel categories while maintaining somewhat the prediction of base categories. 
The combination of RAL and RAF, \ie, RALF, shows significant performance gain of the baselines compared with the sole use of each module. 
To sum up, RAL and RAF demonstrate enhanced performance individually, and their combined shows superior generalization capabilities.

\subsection{Analysis on RALF}
\noindent\textbf{Easy and hard negatives.}
We evaluate the effects of various approaches to handle easy and hard negatives.
First, we use only one type of negative; `Easy negative only' and `Hard negative only'.
Second, we merge the easy and hard negatives into one negative group denoted as `Merged'.
Unlike RAL, $\mathcal{L}^\text{easy}$ of \cref{eq:loss_easy} that uses two types of negatives is not applied in the `Merged' setting.
\Cref{tab:cr_analysis_negatives} shows that the baselines above show inferior performance to our method \textbf{RAL}.
Overall, our method that differentially treats easy and hard negatives boosts performance the most. 
\begin{table}[h]
  \centering
      \begin{tabular}{l|ccc}
        \toprule
         & AP$^\text{N}_{50}$ & AP$^\text{B}_{50}$ & AP$_{50}$ \\
        \midrule
        Easy negative only & 31.0 & \color{gray}{54.3} & \color{gray}{48.2} \\
        Hard negative only & 30.9 & \color{gray}{54.6} & \color{gray}{48.4} \\
        Merged & 30.9 & \color{gray}{55.5} & \color{gray}{49.0} \\
        \midrule
        RAL & 31.3 & \color{gray}{54.5} & \color{gray}{48.4} \\
        \bottomrule
      \end{tabular}
  \caption{\textbf{Analysis of easy and hard negatives on COCO.}}
  \label{tab:cr_analysis_negatives}
\end{table}

\subsection{Qualitative results}
We visualize the detection results to verify that RALF captures novel categories well. 
We compare one of our baselines (\ie, OADP) and OADP + RALF at the top and bottom of \Cref{fig:qualitative}, respectively. 
Each box in the image represents the box prediction results. 
From the qualitative results, RALF captures novel categories better than baseline, which means RALF improves generalizability. 

\section{Conclusion}
In this paper, we present Retrieval-Augmented Losses and visual Features (RALF) that retrieves information from a large vocabulary set and augments losses and visual features. 
To optimize the detector, we add Retrieval-Augmented Losses (RAL), which brings hard and easy negative vocabulary from the pre-defined vocabulary store and reflects the semantic similarity with the ground-truth label. 
Additionally, Retrieval-Augmented visual Features (RAF) augments visual features with generated concepts from a large language model and enables improved generalizability. 
To sum up, RALF combines both modules and easily plugs into various detectors and significantly improves detection ability not only base categories but also novel categories.

\paragraph{Acknowledgements.}
This work was partly supported by ICT Creative Consilience Program through the Institute of Information \& Communications Technology Planning \& Evaluation (IITP) grant (IITP-2024-2020-0-01819), Convergence security core talent training business (Korea University) grant (No.2022-0-01198), and the National Research Foundation of Korea (NRF) grant (NRF-2023R1A2C2005373) funded by the Korea government (MSIT).

%%%%%%%%% REFERENCES
{\small
\bibliographystyle{ieee_fullname}
\bibliography{main}
}

\clearpage
\setcounter{table}{0}
\setcounter{figure}{0}
\renewcommand{\thefigure}{A\arabic{figure}}
\renewcommand{\thetable}{A\arabic{table}}
\renewcommand{\thesection}{A\arabic{section}}
\appendix
\section*{\Large Appendix}
\section{Implementation details}
In this section, we provide hyperparameter settings and implementation details that are not mentioned in the main paper, for each RAL and RAF.

\subsection{RAL details}
After applying the rank variance sampling scheme, we set $m$ to 2,000 and $n$ to 10 to separate the vocabularies into hard and easy negatives.
Different hyperparameters depending on the baseline and the dataset for RAL are shown in \Cref{tab:cr_hyperparameters}.
\begin{table}[h]
    \centering
    \scalebox{1.0}{
        \begin{tabular}{l l | p{.7em} p{.7em} p{.7em} p{.7em} p{.7em} p{.7em}}
        \toprule
        Method & Dataset & $\alpha^\text{hard}$ & $\alpha^\text{easy}$ & $\lambda^\text{hard}$ & $\lambda^\text{easy}$ & $\beta^\text{hard}$ & $\beta^\text{easy}$ \\
        \midrule
        \multirow{2}{*}{OADP~\cite{wang2023object}} & COCO & 0 & 1 & 1 & 5 & 1 & 1 \\
         & LVIS & 0 & 1 & 1 & 10 & 1 & 1 \\
        \midrule
        \multirow{2}{6.3em}{Object-Centric-OVD~\cite{Hanoona2022Bridging}} & COCO & 0 & 1 & 1 & 10 & 1 & 1 \\
         & LVIS & 0 & 1 & 1 & 5 & 1 & 1 \\
        \midrule
        DetPro~\cite{du2022learning} & LVIS  & 0 & 1 & 1 & 10 & 1 & 1 \\
        \bottomrule
        \end{tabular}
    }
    \caption{\textbf{Hyperparameters in RAL.}}
    \label{tab:cr_hyperparameters}
\end{table}

\subsection{RAF details}
When we bring the related concepts from the concept store, we set the number of concepts $k$ to 50.
In the augmenter $\mathcal{A}$, the number of the decoder layers $L$ is set to 6.
The one decoder layer consists of cross-attention (CA) with 8 heads and FFN with 2,048 dimensions.
Positional embeddings $E^\text{pos}$ and type embeddings $E^\text{type}_1$ and $E^\text{type}_2$ are initialized with random values.
The total number of parameters for the augmenter $\mathcal{A}$ is 51M.
During RAF training, we use $\beta^\text{cls}$ of 5.0 and $\beta^\text{reg}$ of 1.0.

\section{Further ablation study}
\subsection{RAF in official baseline}
The performance figures of the baseline compared to RALF in the main paper Section 4.2 are the results we reproduced ourselves for the fairness of the ablations.
We checked the potential variation in performance depending on the execution environment. 
Therefore, we further verified how the performance changes when RAF is applied to the officially provided model checkpoints by the baseline authors.
As depicted in \Cref{tab:cr_raf_ablation_coco}, RAF demonstrates significant performance improvements on the COCO datasets.
\begin{table}[h]
  \centering
  \scalebox{0.95}{
      \begin{tabular}{l|c|cccc}
        \toprule
        Method & RAF & AP$^\text{N}_{50}$ & AP$^\text{B}_{50}$ & AP$_{50}$ \\
        \midrule
        \multirow{2}{*}{OADP~\cite{wang2023object}} & & 31.3 & \color{gray}{55.0} & \color{gray}{48.8} \\
        & \checkmark & 33.1 & \color{gray}{54.7} & \color{gray}{49.0} \\
        \midrule
        \multirow{2}{*}{Object-Centric-OVD~\cite{Hanoona2022Bridging}} & & 40.7 & \color{gray}{54.1} & \color{gray}{50.6} \\
         & \checkmark & 41.1 & \color{gray}{54.2} & \color{gray}{50.8} \\
        \bottomrule
      \end{tabular}
  }
  \caption{\textbf{RAF ablation in the official checkpoint on COCO.}}
  \label{tab:cr_raf_ablation_coco}
\end{table}

\section{Further analysis}
\begin{figure}[h]
  \centering
  \includegraphics[width=1.0\linewidth]{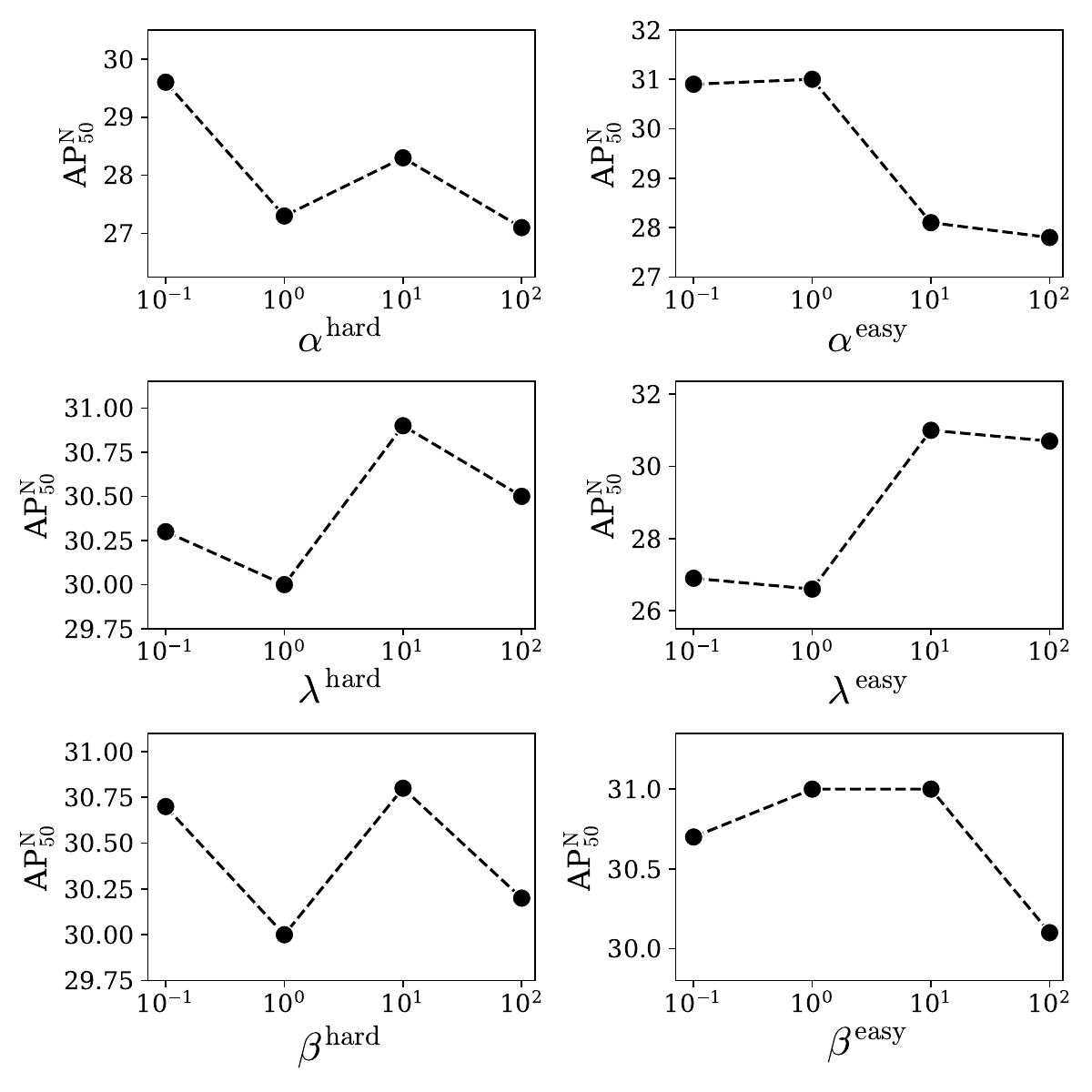}
   \caption{\textbf{Analysis on hyperparameters of $\mathcal{L}^\text{RAL}$.}}
   \label{fig:cr_hyperparameters}
\end{figure}

\noindent\textbf{Hyperparameters of $\mathcal{L}^\text{RAL}$.}
As depicted in~\Cref{fig:cr_hyperparameters}, we provide an analysis of the hyperparameters controlling $\mathcal{L}^\text{RAL}$ in OADP~\cite{wang2023object} on COCO dataset. To conduct the analysis, we fixed $\alpha^\text{hard}$, $\lambda^\text{hard}$, and $\beta^\text{hard}$ as (0, 1, 1) and $\alpha^\text{easy}$, $\lambda^\text{easy}$, and $\beta^\text{easy}$ as (1, 10, 1), respectively. 
The analysis is performed using values (0.1, 1, 10, 100) for 6 hyperparameters.
When increasing $\alpha^\text{hard}$ from 1 to 10, we observed a slight improvement, however, there was a declining performance tendency overall.
$\alpha^\text{easy}$ exhibited a temporary performance increase at 1 followed by a subsequent decline. 
Relative to lower values, $\lambda^\text{easy}$ demonstrated better performance with higher values.
Regarding $\lambda^\text{hard}$, $\beta^\text{hard}$, and $\beta^\text{easy}$, we found that they exhibited robust performance even with varying values.

\noindent\textbf{Exploration of $n$.}
\Cref{tab:cr_no_random_n} shows the difference between obtaining $n$ of $m$ negative vocabularies for each iteration based on random or similarity in RAL.
The similarity performance is better than the baseline in $\text{AP}_{50}^{\text{B}}$ and $\text{AP}_{50}$, however, it was observed that the performance was lower in $\text{AP}_{50}^{\text{N}}$ due to the absence of randomness. Adopting randomly selecting $n$ of $m$ negative vocabularies showed the best performance for novel categories.
\begin{table}[h]
    \centering
    \scalebox{1.0}{
        \begin{tabular}{l | c c c}
        \toprule
        Method & AP$^\text{N}_{50}$ & AP$^\text{B}_{50}$ & AP$_{50}$ \\
        \midrule
        Baseline & 30.0 & \color{gray}{54.5} & \color{gray}{48.1} \\
        Similarity & 29.7 & \color{gray}{54.9} & \color{gray}{48.3} \\
        Random & \textbf{31.3} & \color{gray}{54.5} & \color{gray}{48.4} \\
        \bottomrule
        \end{tabular}
    }
    \caption{\textbf{Comparison on obtaining $n$.}}
    \label{tab:cr_no_random_n}
\end{table}

\noindent\textbf{Sampling schemes.}
\label{abl:sampling_scheme}
As described in Section 3.3 of the main paper, RAL defines hard and easy negative vocabularies through rank variance sampling of negative retrievers from the vocabulary store. 
To ascertain the optimality of the rank variance sampling method, we compared various sampling schemes -- random sampling and similarity-based sampling, and these results are reported in ~\Cref{tab:cr_scheme_coco}. 
Random sampling scheme extracts vocabularies randomly without considering other factors, while similarity-based sampling scheme reflects the cosine similarity between $C^\mathcal{V}$ and ground-truth labels. 
The experimental results show that among the various sampling schemes, rank variance sampling not only demonstrates superior performance in $\text{AP}_\text{50}^{\text{N}}$but also exhibits outstanding performance in $\text{AP}_\text{50}$. 
\begin{table}[h]
  \centering
  \scalebox{1.0}{
      \begin{tabular}{l|ccc}
        \toprule
        Sampling scheme & AP$^\text{N}_{50}$ & AP$^\text{B}_{50}$ & AP$_{50}$ \\
        \midrule
        Random & 30.6 & \color{gray}{54.4} & \color{gray}{48.1} \\
        Similarity-based & 30.2 & \color{gray}{54.8} & \color{gray}{48.3} \\
        Rank variance & \textbf{31.3} & \color{gray}{54.5} & \color{gray}{48.4} \\
        \bottomrule
      \end{tabular}
  }
  \caption{\textbf{Analysis on sampling scheme.}}
  \label{tab:cr_scheme_coco}
\end{table}

\noindent\textbf{Hyperparameter $k$ in RAF.}
\Cref{tab:cr_topk_concept_coco} and~\Cref{tab:cr_topk_concept_lvis} show how much performance varies depending on the $k$ used to augment visual features in RAF on COCO and LVIS dataset, respectively.
We set $k$ to 50 as the default.
When $k=10$, the best performance was observed on COCO. 
The minor performance changes are exhibited across different values of $k$ on LVIS.
The results showed that our method is robust to the choice of $k$.
\begin{table}[h]
    \centering
    \scalebox{1.0}{
        \begin{tabular}{l | c c c}
        \toprule
        $k$ & AP$^\text{N}_{50}$ & AP$^\text{B}_{50}$ & AP$_{50}$ \\
        \midrule
        10 & \textbf{33.6} & \color{gray}{54.4} & \color{gray}{49.0} \\
        20 & 33.3 & \color{gray}{54.4} & \color{gray}{48.9} \\
        50 & 33.4 & \color{gray}{54.5} & \color{gray}{49.0} \\
        100 & 33.3 & \color{gray}{54.4} & \color{gray}{48.9} \\
        \bottomrule
        \end{tabular}
    }
    \caption{\textbf{Analysis on $k$ in RAF on COCO.}}
    \label{tab:cr_topk_concept_coco}
\end{table}

\begin{table}[h]
    \centering
    \scalebox{1.0}{
        \begin{tabular}{l | c c c c}
        \toprule
        $k$ & AP$_\text{r}$ & AP$_\text{c}$ & AP$_\text{f}$ & AP \\
        \midrule
        10 & \textbf{21.9} & \color{gray}{26.2} & \color{gray}{29.1} & \color{gray}{26.6} \\
        20 & \textbf{21.9} & \color{gray}{26.2} & \color{gray}{29.1} & \color{gray}{26.6} \\
        50 & \textbf{21.9} & \color{gray}{26.2} & \color{gray}{29.1} & \color{gray}{26.6} \\
        100 & 21.8 & \color{gray}{26.2} & \color{gray}{29.1} & \color{gray}{26.6} \\
        \bottomrule
        \end{tabular}
    }
    \caption{\textbf{Analysis on $k$ in RAF on LVIS.}}
    \label{tab:cr_topk_concept_lvis}
\end{table}

\noindent\textbf{Scale of large vocabulary.}
As discussed in Section 3.3 of the main paper, RALF uses a large vocabulary set to construct the vocabulary store. 
We set the experiment to analyze the effectiveness of the vocabulary size. 
We adopt V3Det~\cite{wang2023v3det} with 13,204 vocabularies as the vocabulary set for all experiments. 
Before retrieving hard and easy negatives from the vocabulary store, we construct the vocabulary store by eliminating unnecessary elements, \ie, novel categories and case-overlapping, from the vocabulary set. 
Following this process, we obtained 13,064 vocabularies from V3Det. 
To examine whether vocabulary size affects performance, we experiment on COCO benchmarks by reducing the percentage of vocabulary size to 40\% and 70\%. 
The results are shown in ~\Cref{tab:cr_vocabscale_coco}. 
From the results, we observed that the performance was better when the vocabulary size was 100\% compared to when it was 40\% or 70\%.
\begin{table}[h]
  \centering
  \scalebox{1.0}{
      \begin{tabular}{c|ccc}
        \toprule
        Scale of large vocabulary & AP$^\text{N}_{50}$ & AP$^\text{B}_{50}$ & AP$_{50}$ \\
        \midrule
        40\% & 30.9 & \color{gray}{54.7} & \color{gray}{48.5} \\
        70\% & 30.8 & \color{gray}{54.7} & \color{gray}{48.4} \\
        100\% & \textbf{31.3} & \color{gray}{54.5} & \color{gray}{48.4} \\
        \bottomrule
      \end{tabular}
    }
  \caption{\textbf{Analysis on a large vocabulary size.}}
  \label{tab:cr_vocabscale_coco}
\end{table}

\noindent\textbf{Retrieve negative vocabularies with BERT.}
In this work, the hard and easy negative vocabularies are retrieved based on cosine similarity between CLIP~\cite{radford2021learning} text embeddings in RAF. 
When retrieving negative vocabularies, it is also possible to use the embeddings of a language model (LM) instead of CLIP. 
We extract embeddings about base categories and a large vocabulary set with a language model BERT~\cite{devlin2019bert} and then retrieve hard and easy negative vocabularies based on the cosine similarity between the embeddings.
\Cref{tab:cr_bert} depicts the comparison results of CLIP and BERT.
The performance on the novel categories is lower in BERT, but it increases by 0.7 AP$^\text{N}_{50}$ compared to the baseline.
\begin{table}[h]
    \centering
    \scalebox{1.0}{
        \begin{tabular}{l | c c c}
        \toprule
        Method & AP$^\text{N}_{50}$ & AP$^\text{B}_{50}$ & AP$_{50}$ \\
        \midrule
        Baseline & 30.0 & \color{gray}{54.5} & \color{gray}{48.1} \\
        CLIP~\cite{radford2021learning} & \textbf{31.3} & \color{gray}{54.5} & \color{gray}{48.4} \\
        BERT~\cite{devlin2019bert} & 30.7 & \color{gray}{55.4} & \color{gray}{48.9} \\
        \bottomrule
        \end{tabular}
    }
    \caption{\textbf{CLIP vs BERT.}}
    \label{tab:cr_bert}
\end{table}

\end{document}